\relax
\documentclass[letterpaper]{article} 
\usepackage[accepted] {icml2019} 
\usepackage{times}  
\usepackage{helvet}  
\usepackage{courier}  
\usepackage{url}  
\usepackage{graphicx}  
\usepackage{amsmath}
\usepackage{mathrsfs}
\usepackage{amsfonts}
\usepackage{subfig}
\usepackage{array}
\usepackage{booktabs} 
\usepackage{bm}
\usepackage{hyperref} 
\usepackage[small,compact]{titlesec}
\usepackage{combelow}
\newcommand{\topic}[1]{\vspace{0.01in}\noindent\textbf{#1.}} 
\newcommand{\nstopic}[1]{\noindent\textbf{#1.}} 

\newcommand\arch[1]{\bm{$\mathcal{#1}$}}

\newcolumntype{?}[1]{!{\ width #1}}

\begin{document}

\twocolumn[
\icmltitle{Shallow-Deep Networks: Understanding and Mitigating Network Overthinking}



\icmlsetsymbol{equal}{*}

\begin{icmlauthorlist}
\icmlauthor{Yigitcan Kaya}{to}
\icmlauthor{Sanghyun Hong}{to}
\icmlauthor{Tudor Dumitra\cb{s}}{to}

\end{icmlauthorlist}

\icmlaffiliation{to}{University of Maryland, Maryland, USA}

\icmlcorrespondingauthor{Yigitcan Kaya}{cankaya@umiacs.umd.edu}

\icmlkeywords{Machine Learning, Deep Learning}

\vskip 0.3in
]



\printAffiliationsAndNotice{}  

%
%

\begin{abstract}
We characterize a prevalent weakness of deep neural networks (DNNs)---\emph{overthinking}---which occurs when a DNN can reach correct predictions before its final layer.
Overthinking is computationally \emph{wasteful}, and it can also be \emph{destructive} when, by the final layer, a correct prediction changes into a misclassification.
%
%
Understanding overthinking requires studying how each prediction \emph{evolves} during a DNN's forward pass, which conventionally is opaque.
%
%
For prediction transparency, we propose the Shallow-Deep Network (SDN), a generic modification to \emph{off-the-shelf DNNs} that introduces \emph{internal classifiers}. 
We apply SDN to four modern architectures, trained on three image classification tasks, to characterize the overthinking problem.
%
%
We show that SDNs can mitigate the wasteful effect of overthinking with confidence-based \emph{early exits}, which reduce the average inference cost by more than 50\% and preserve the accuracy.
We also find that the destructive effect occurs for 50\% of misclassifications on natural inputs and that it can be induced, adversarially, with a recent backdooring attack. 
To mitigate this effect, we propose a new \emph{confusion} metric to quantify the internal disagreements that will likely lead to misclassifications. 

\end{abstract}
%
%

\section{Introduction}
\label{sec:intro}
\newcommand{\fnOne}{Prior work has utilized this term to describe a property of the network as a whole, disregarding its internal state~\cite{wang2017idk}. In contrast, our definition is closely tied to the network's internal state and is analogous to human overthinking.}
\newcommand{\fnTwo}{\href{http://tiny-imagenet.herokuapp.com/}{http://tiny-imagenet.herokuapp.com/}}
\newcommand{\fnThree}{\href{www.shallowdeep.network}{www.shallowdeep.network}}

Deep neural networks (DNNs) have enabled breakthroughs in many tasks, such as image classification~\cite{krizhevsky2012imagenet} and speech recognition~\cite{hinton2012deep}.
In these tasks, a DNN's sequence of layers resembles human perception in the way it combines simple representations, such as edges, into more complex ones, such as faces~\cite{zeiler2014visualizing}.
A fundamental difference is that people can learn simpler heuristics that allow them to perform even complex tasks, such as driving or playing the piano, with little mental effort~\cite{Kahneman11:ThinkingFastSlow}.
When simpler heuristics are adequate to complete the task, using excessively complex representations leads to \emph{overthinking}.
\emph{Human overthinking} is considered wasteful because it leads to slow decision making, as people think too much or for too long.
Furthermore, overthinking is potentially destructive, by causing confusion and mistakes when human attention is drawn to irrelevant details.

%
In contrast, the decision-making process of conventional DNNs---the forward pass---requires the same computational effort on all inputs, whether they are simple or difficult to classify.
Building on this analogy, we ask: \emph{Are deep neural networks also susceptible to overthinking?}.
We consider that a network overthinks\footnote{\fnOne} on an input sample when its simpler representations at an earlier layer---relative to the final layer--- are adequate to make a correct classification. 
Analogously to human overthinking, we hypothesize that further computation after this layer leads to waste and, potentially, a misclassification.

Our definition of overthinking relates to how a prediction \emph{evolves} throughout the forward pass.
Intuitively, a DNN produces predictions through a gradual process, as the subsequent layers recognize different features of the input.
In conventional DNNs, however, this process remains mostly opaque as they are only able to provide a \textit{final prediction}.
The prior work that proposed early exits~\cite{lee2015deeply, szegedy2015going} or error indicators~\cite{szegedy2013intriguing, papernot2018deep} for off-the-shelf DNNs focused on producing and interpreting a single prediction for each input, rather than on illuminating how this prediction evolves from layer to layer.

\begin{figure}
    \centering
    \includegraphics[width=0.48\textwidth]{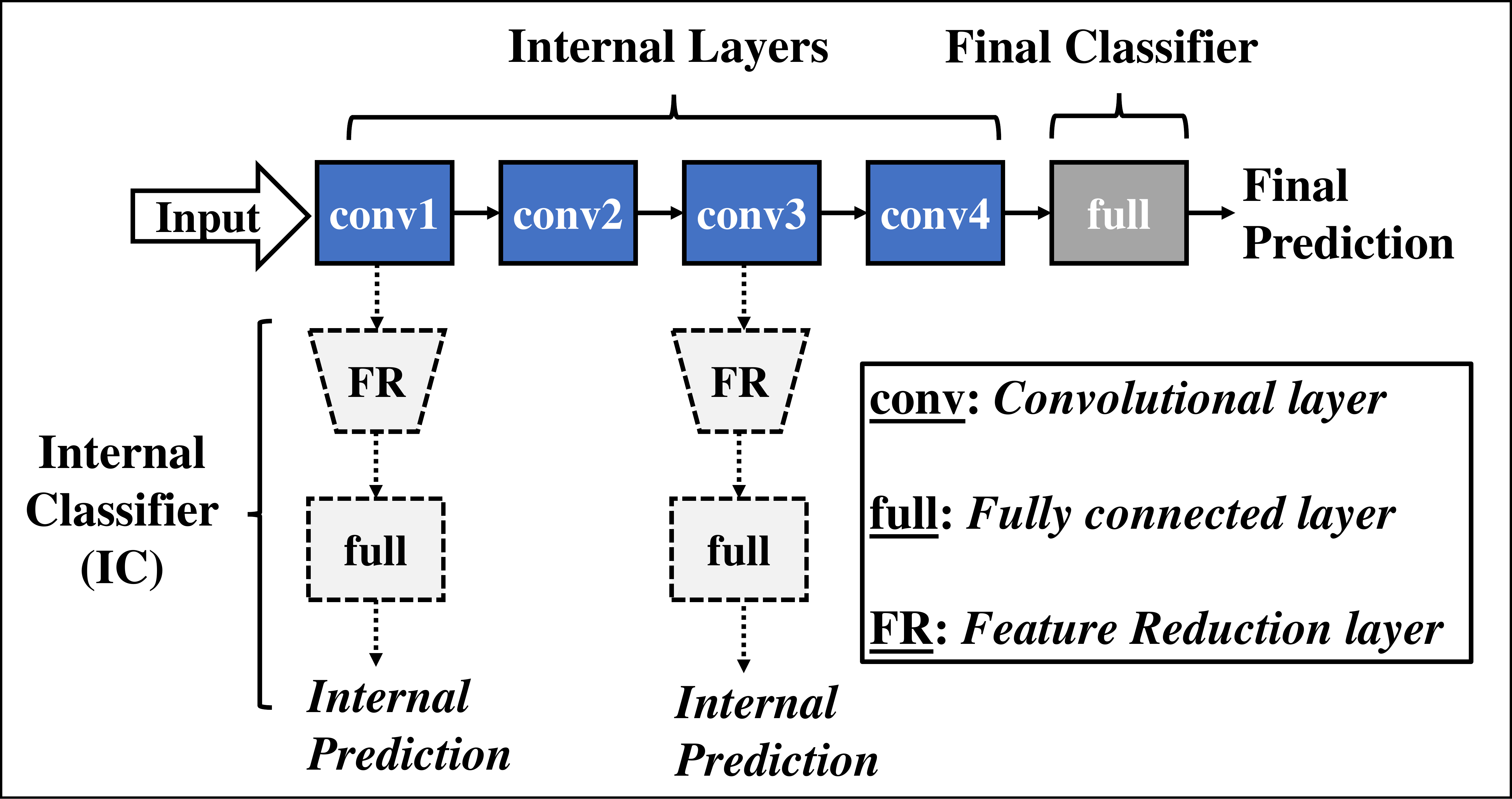}
    \caption{Modifying a standard convolutional neural network as a Shallow-Deep Network. Our modification introduces the dotted components---two ICs. The resulting network produces three predictions: two internal and one final.}
    \label{fig:sdn_arch}
\end{figure}

Our first contribution is the \emph{Shallow-Deep Network} (SDN): a generic modification to off-the-shelf DNNs for introducing \emph{internal classifiers} (ICs).
Our modification attaches ICs to various stages of the forward pass, as Figure~\ref{fig:sdn_arch} illustrates, and effectively combines shallower and deeper networks into one.
The feature reduction in an IC acts as a regularizer and ensures that our method can scale up to large DNNs.
We can apply the modification to both pre-trained and untrained DNNs.
The conversion from a pre-trained DNN is efficient as we only train the parameters in the ICs.
Moreover, by using a weighted loss function, we can also train the original network from scratch jointly with the ICs.
This mode of training, the \emph{SDN training}, often improves the original accuracy and yields more accurate ICs.
Our modification is practical for a range of existing architectures and allows us to explore the overthinking problem.

\raggedbottom

Our second contribution is to show that convolutional neural networks (CNNs) overthink on the majority of inputs.
Our study exposes that CNN overthinking can be both \emph{wasteful} and \emph{destructive}.
When a CNN reaches a correct internal prediction before its final layer, the remaining layers are effectively rendered redundant.
Our experiments show that, for up to $\sim$95\% of instances, overthinking leads to slow inferences and wasted computation.
This suggests that the \emph{complex} inputs requiring the full network depth are uncommon.
More surprisingly, we observe the destructive effect of overthinking in up to $\sim$50\% of a CNN's errors on natural inputs, where a correct internal prediction evolves into a misclassification.
Moreover, we show that a recent backdooring attack on CNNs~\cite{gu2017badnets} induces the same effect on the victim network.

We further utilize SDNs as a vehicle for mitigating the overthinking problem.
%
%
As in human problem-solving, we cannot evaluate perfectly whether a classification is correct, but we can utilize heuristics against overthinking. 
%
%
Our third contribution is to propose two different heuristics: confidence-based \emph{early exits} and analysis of \emph{confusion}.

Our first heuristic uses the \emph{confidence} of an internal prediction to assess its correctness.
With this heuristic, we can reliably detect when the network should stop \emph{thinking} and make an early prediction---an \emph{early exit}.
Without any loss of accuracy, we can reduce the average inference cost by up to $\sim$75\% and mitigate the wasteful effect of overthinking.
%
The destructive effect of overthinking, on the other hand, poses a greater challenge for an early exit mechanism.
The disagreements among the internal predictions hint the state of \textit{confusion} the network is in.
The confusion hurts the confidence of correct internal predictions and, as a result, leads inputs to bypass the early exits. 
In our second heuristic, we devise a new \emph{confusion metric} that quantifies the internal disagreements.
We observe experimentally that confusion scores reliably indicate whether the network is likely to misclassify an input.
Furthermore, by visualizing the confusion---the disagreements---we can investigate the input elements that cause the confusion.
In addition to their practical application regarding diagnosing DNN errors, these visualizations provide a new perspective for reasoning about model interpretability.

We evaluate our techniques on three tasks: CIFAR-10, CIFAR-100~\cite{krizhevsky2009learning}, and Tiny ImageNet\footnote{\fnTwo}; by applying the SDN modification to four off-the-shelf CNN architectures: VGG~\cite{DBLP:journals/corr/SimonyanZ14a}, ResNet~\cite{he2016deep}, Wide ResNet~\cite{zagoruyko2016wide} and MobileNets~\cite{DBLP:journals/corr/HowardZCKWWAA17}.
An SDN's early exits mitigate the wasteful effect of overthinking and cut the average inference costs by more than 50\% in CIFAR-10 and CIFAR-100, and by more than 25\% in Tiny ImageNet.
Further, early exits can improve a CNN's accuracy by up to 8\% and recover the accuracy of a backdoored CNN from 12\% to 84\% on malicious inputs.
Moreover, our normalized confusion scores suggest that a network is significantly less confused in a correct prediction ($-0.3$ on average) than in a wrong prediction ($0.7$ on average).
We also release all of our source code\footnote{\fnThree}.


%
%

\section{Preliminaries}
\label{sec:prelim}
\subsection{Related Work}
\nstopic{Internal classifiers \& Early exits}
Deeply Supervised Networks~\cite{lee2015deeply} and Inception architecture~\cite{szegedy2015going} have proposed using internal classifiers for improving overall accuracy.
In~\cite{pmlr-v80-huang18b} and~\cite{DBLP:journals/corr/abs-1812-11446}, authors propose methods to train a DNN sequentially, block-by-block, for reducing the training costs.
These studies discard the internal classifiers, as they are not designed, or trained, for accurate predictions.
The MSDNets~\cite{huang2017multi} architecture aims to provide any time predictions, in the case of insufficient computing resources.
Here, the authors claim that attaching internal classifiers to existing architectures hurts the final performance.
BranchyNets~\cite{teerapittayanon2016branchynet} architecture augments DNNs for improving the inference times, but it requires training the network from scratch.
Adaptive Neural Networks~\cite{bolukbasi2017adaptive} and Runtime Neural Pruning~\cite{lin2017runtime} intend to reduce the evaluation time by selectively evaluating the components of a network.
All these methods implicitly attempt to mitigate only the wasteful effect of overthinking; however, our primary goal is to fully understand then remedy the problem.
Further, we design Shallow-Deep Networks as a modification, instead of a new architecture, for introducing internal classifiers to pre-trained off-the-shelf networks.
Our modification achieves two goals previous studies have suggested to be contradictory: having accurate internal classifiers while preserving the original performance.

\topic{Neural Network Error Indicators}
Widespread usage of DNNs makes indications of prediction correctness crucial for handling the potential errors.
Prior work has shown that predicted probability distribution---\textit{softmax scores}---makes an unreliable metric, as it is over-confident towards a single class~\cite{szegedy2013intriguing}.
In this regard, prior work has described a calibration scheme~\cite{guo2017calibration} and Bayesian model uncertainty estimation~\cite{gal2016dropout}.
The credibility metric~\cite{papernot2018deep} uses the representational distances to neighboring training points.
SDNs inherently allow the design of our \emph{confusion} metric that does not require any calibration or expensive computations, and can be applied to pre-trained off-the-shelf DNNs.
Confusion captures how consistently a network reaches a final prediction.
Further, our method enables diagnosis of the errors by visualizing the input elements responsible for the confusion.

\subsection{Experimental Setup}
\nstopic{Datasets}
In our experiments, we use three datasets for benchmarking: CIFAR-10, CIFAR-100~\cite{krizhevsky2009learning} and Tiny ImageNet.
The two CIFAR datasets consist of 32x32 pixels, colored natural images. 
CIFAR-10 and CIFAR-100 images are drawn from 10 and 100 classes, respectively; containing 50,000 training and 10,000 validation images. 
We use a standard data augmentation scheme: padding, random cropping, and random horizontal mirroring.
The Tiny ImageNet dataset consists of a subset of ImageNet images~\cite{deng2009imagenet}, resized at 64x64 pixels.
There are 200 classes, each of which has 500 training and 50 validation images.
We augment the data with random crops, horizontal mirroring, and RGB intensity scaling.

\topic{Architectures and Hyper-Parameters}
We experiment with four off-the-shelf CNNs: VGG~\cite{DBLP:journals/corr/SimonyanZ14a}, ResNet~\cite{he2016deep}, Wide-ResNet (WRN)~\cite{zagoruyko2016wide} and MobileNet~\cite{DBLP:journals/corr/HowardZCKWWAA17}.
Specifically, we use VGG-16, ResNet-56 and WRN-32-4 configurations of these architectures.
We train the CNNs for 100 epochs, using the hyper-parameters the original studies describe.
In our experiments, we use these CNNs as our pre-trained, off-the-shelf networks.
We denote the network after our SDN modification by appending \textit{`SDN'} to its original name; e.g. \textit{VGG-16} becomes \textit{VGG-16-SDN}.
To apply SDNs to pre-trained networks, we train the internal classifiers for 25 epochs, using the Adam optimizer~\cite{kingma2014adam}.
If we start training a modified network from scratch, we train for 100 epochs; the same as the original networks.

\topic{Metrics}
To quantify the test-time inference cost of a network, we measure the average number of floating point operations (\textit{FLOPs}) a network performs to classify an input.
As for the classification performance, we simply report the Top-1 accuracy on the test data (as a percentage).

%
%

\section{The Shallow-Deep Network}
\label{sec:shallow_deep}
\nstopic{Setting}
We consider the supervised learning setting and the standard DNN structure: a sequence of \textit{internal layers} ending with a \textit{final classifier} (see Figure~\ref{fig:sdn_arch}). 
Let $S = (x, y) \in (X,Y)$ be our input-output pairs, where $x$ denotes a sample, and $y \in \{1,...K\}$ is its correct ground-truth label.
Given $x$, a DNN with $M$ internal layers performs a \textit{classification}, $\mathcal{F}_{final}(F_M(,...F_1(x)))$, to predict the probability of $x$ belonging to each class label. 
Here, $F_m$ denotes the \textit{learnable} function layer $m$ applies, and $\mathcal{F}_{final}$ denotes the final classifier---usually a fully connected layer.
To simplify the notation, we write the output of the layer $m$ as $F_m(x)$ and the vector of predicted probabilities as $\mathcal{F}_{final}(x)$.
We denote the \textit{final prediction} on the sample $x$ as $\hat{y}_{final}=\operatorname*{argmax}_j \mathcal{F}_{final}^{(j)}(x)$, i.e. the predicted label with the highest probability.

\topic{Overview}
In this work, we specifically focus on off-the-shelf CNNs and leave the extension to other architectures for future work.
Figure~\ref{fig:sdn_arch} outlines how we apply our modification.
We call a modified network a Shallow-Deep Network (SDN).
An SDN contains \emph{internal classifiers} (ICs) and essentially combines \emph{shallow} networks (earlier ICs) and \emph{deep} networks (later ICs and the final classifier) into a single structure. 
In addition to the final prediction $\hat{y}_{final}$, an SDN produces multiple \emph{internal predictions} at its ICs.
Formally, $i^{th}$ IC ($IC_i$) following the layer $m$ is a learnable function: $\mathcal{F}_{i,m}$, $1\leq m \leq M$ and $1 \leq i \leq N$ ($N$ is the total number of ICs).
$\mathcal{F}_{i,m}$ takes $F_m(x)$ as its input and performs a classification on the sample $x$: $\mathcal{F}_{i,m}(F_m(x))$, or simply as $\mathcal{F}_i(x)$.
We denote the $i^{th}$ internal prediction as $\hat{y}_i = \operatorname*{argmax}_j \mathcal{F}_i^{(j)}(x)$.
The modification includes two steps: attaching several ICs (Section~\ref{ssec:attach}), and training these ICs (Section~\ref{ssec:train}). 
We elaborate on the cost of SDNs in Section~\ref{ssec:cost}.
We demonstrate the outcome of such modification and the accuracy of the ICs in Section~\ref{ssec:ic_accuracy}.

\subsection{Attaching the Internal Classifiers (ICs)}
\label{ssec:attach}
An internal classifier consists of two parts: a single fully connected layer that follows a feature reduction layer.
Large output sizes of a CNN's internal layers necessitate the feature reduction for scalability.
The feature reduction layer takes $F_m(x)$ and reduces its size.
The fully connected layer, using the reduced $F_m(x)$, produces the internal prediction.
Instead of using a more complex structure, such as convolutional layers or a multilayer perceptron~\cite{szegedy2015going,teerapittayanon2016branchynet}, we opt for using a fully connected layer to keep our modification minimal and generic.
Further, this allows us to closely monitor how predictions evolve throughout the forward pass in Section~\ref{sec:overthinking} and highlight the prevalance of the overthinking problem without any interference from our IC mechanism.

\topic{Feature Reduction Layers}
We observe that neither average nor max pooling consistently performs better for reducing the feature map size---the output dimension of one convolution filter.
Here, we employ a \textit{mixed} max-average pooling strategy~\cite{lee2016generalizing}, which learns the mixing proportion of two pooling methods from the data, without any manual tuning.
We simply pick the pooling size such that any feature map size larger than $4$x$4$ is pooled into $4$x$4$---the smaller sizes remain the same.
Without any reduction---with only the fully connected layer---ICs increase the size of the original network by up to $\sim$18x. This is brought down to at most $\sim$3x by feature reduction.

\topic{Placement of the ICs}
We pick a subset of internal layers to attach the internal classifiers after.
Our selection criterion is simple: we pick the internal layers closest to the 15\%, 30\%, 45\%, 60\%, 75\% and 90\% of the full network's inference cost---the number of \emph{FLOPs}.
Given an input, our SDNs produce a total of 6 internal predictions and a final prediction.
For each IC, we denote the \textit{relative inference cost to the whole network} as $\mathcal{C}_i : 1 \leq i \leq N$---for our SDNs, $N=6$ and $\mathcal{C}_1 \approx 0.15...\mathcal{C}_6 \approx 0.9$.
For the final classifier: $\mathcal{C}_{final} = 1$.
Overall, this simplifies our analysis of overthinking and eliminates the need for tuning.

\subsection{Training the Internal Classifiers}
\label{ssec:train}
To complete our modification, we train the ICs using the training set, via backpropagation.
The training strategy is based on whether the original CNN is pre-trained---the \emph{IC-only training}---, or untrained---the \emph{SDN training}.

\topic{The IC-only Training}
To \textit{convert} a pre-trained network to an SDN, we freeze original weights and train only the weights in the attached ICs.
Freezing prevents any change to the off-the-shelf network, and, as a result, keeps its predictions fully intact.
The IC-only training is fast as the backpropagation updates only the weights in the ICs.

\topic{The SDN Training}
The IC-only training has a major drawback: the standard network training aims to improve only the final prediction accuracy.
Attaching classifiers on top of these pre-trained internal layers, as a result, yields relatively weak ICs.
Here, we propose a \emph{weighted loss function} to train the original weights, from scratch, jointly with the ICs, similar to previous studies~\cite{lee2015deeply}.
However, unlike previous methods, we design an objective function for achieving high accuracy in all ICs in addition to the final classifier.
Our objective multiplies each $\mathcal{L}_i$, the training loss values from $i^{th}$ IC, with coefficients $\tau_i$. It then adds their sum to the loss from the final classifier.
Our selection of $\tau_i$ reflects two observations: (i) the initial phases of the training are less stable due to higher learning rate; and (ii) the earlier ICs have less learning capacity.
In this regard, we start the training with $\tau_i=0.01$ and linearly increase it until $\tau_i = \mathcal{C}_i$---the $i^{th}$ IC's relative inference cost.

\subsection{Cost of the Modification}
\label{ssec:cost}
Table~\ref{table:sdn_cost} highlights the cost of attaching ICs on Tiny ImageNet task.
Here, we do not consider any early exits at all; and the input samples traverse until the end of the SDN.
The overhead on the required test-time computation---FLOPs per input---is minor; as the fully connected layers are computationally efficient.
As expected, the IC-only training is $\sim$3x faster than original training---it also requires $\sim$4x fewer training epochs.
In the SDN training strategy, on the other hand, we see a 10\% increase in training time.
Finally, because of the fully connected layers, the modification increases the number of parameters. 

\begin{table}
    \caption{Computational comparison between off-the-shelf CNNs and SDNs. The left side of a cell is for the CNNs. \textbf{\#prms} presents the total number of parameters. \textbf{\#ops} reports the total number of FLOPs performed for each sample. \textbf{epc} lists the training time per epoch---For SDNs both the IC-only and the SDN training strategies are listed.}
    \label{table:sdn_cost}
    \centering
    \begin{small}
    \begin{sc}
    \begin{tabular}{lccr}
        \toprule
        \textbf{Network} &\textbf{\#prms} (M) &\textbf{\#ops} (B) &\textbf{epc} (min)\\
        \midrule
        \small{\arch{V}GG-16}&\small{21.2 $|$ 26.8}&\small{2.5 $|$ 2.5}&\small{3.4 $|$ 1.4 - 3.8}\\
        \small{\arch{R}esNet-56}&\small{0.9 $|$ 1.6}&\small{0.2 $|$ 0.2}&\small{1.0 $|$ 0.5 - 1.1}\\
        \small{\arch{W}RN-32-4}&\small{7.4 $|$ 10.3}&\small{2.4 $|$ 2.4}&\small{3.5 $|$ 1.2 - 3.9}\\
        \small{\arch{M}obileNet}&\small{3.4 $|$ 11.2}&\small{3.0 $|$ 3.0}&\small{1.9 $|$ 0.8 - 2.1}\\
        \bottomrule
    \end{tabular}
    \end{sc}
    \end{small}
\end{table}
\raggedbottom

\subsection{The Accuracy of the Internal Classifiers}
\label{ssec:ic_accuracy}
Table~\ref{table:internal_classifiers} presents the relative accuracy of the ICs on Tiny ImageNet task.
The SDN training consistently enables ICs that have higher accuracy than the original network.
This improvement is more pronounced on more difficult tasks, increasing the original accuracy by up to 4.5\%.
The SDN training optimizes the internal layers for highly discriminative features, similar to~\cite{lee2015deeply}.
As a result, the classifiers built on these features outperform their counterparts.
Even with the IC-only training, ICs have relatively high accuracy that exceeds the original accuracy in one case.
Further, the networks reach most of their original accuracy at an IC; which has significantly less inference cost.
Altogether, the accuracy of the ICs demonstrates the extent of the computational waste overthinking causes.

\begin{table}
    \caption{The accuracy of the internal classifiers (ICs) on Tiny ImageNet task. \textbf{Network} lists the original CNNs and their test accuracies. \textbf{$\sim$80\% IC} and \textbf{$\sim$90\% IC} list the IC's index (from $1$ to $6$) which has accuracy closest to 80\% and 90\% of the CNN's. \textbf{Max\% IC} presents the highest accuracy of an IC and its index. We highlight when an IC's accuracy exceeds the CNN's. The left and right sides of each cell are for the IC-only and the SDN training strategies.}
    \label{table:internal_classifiers}
    \centering
    \begin{small}
    \begin{sc}
        \begin{tabular}{lccr}
            \toprule
            \textbf{Network}&\textbf{$\sim$80\% IC}&\textbf{$\sim$90\% IC}&\textbf{Max\% IC}\\
            \midrule
            \small{\arch{V}\textit{(58.6)}} & $3$ $|$ $2$ & $4$ $|$ $3$ & \textbf{59.3($6$)} $|$ \textbf{63.1}($6$) \\
            \small{\arch{R}\textit{(53.9)}} & $4$ $|$ $3$ & $4$ $|$ $3$ & 50.1($6$) $|$ \textbf{54.1}($6$) \\
            \small{\arch{W}\textit{(60.3)}} & $5$ $|$ $3$ & $6$ $|$ $4$ & 54.6($6$) $|$ \textbf{62.2}($6$) \\
            \small{\arch{M}\textit{(59.3)}} & $3$ $|$ $2$ & $4$ $|$ $3$ & 58.3($6$) $|$ \textbf{59.6}($6$) \\
            \bottomrule
        \end{tabular}
    \end{sc}
    \end{small}
\end{table}
\raggedbottom
\section{Understanding the Overthinking Problem}
\label{sec:overthinking}
Shallow-Deep Networks, by providing explicit internal predictions, allow us to study the overthinking problem in CNNs.
In the context of an SDN, we define overthinking on an input sample as the network's ability to reach a correct internal prediction.
Formally, the network overthinks on the input sample $(x,y)$ if $\hat{y}_i = y$, i.e. the $i^{th}$ internal prediction is a correct one.

Overthinking is problematic as the rest of the forward pass ends up being invoked for no clear benefit.
The computational waste this leads to is \textit{the wasteful effect} of overthinking (Section~\ref{ssec:wasteful}).
Moreover, in some cases, excessive processing ultimately leads to a misclassification, i.e. $\hat{y}_i = y \neq \hat{y}_{final}$; or causes a vulnerability~\cite{wang2018one}.
We categorize this as \textit{the destructive effect} of overthinking (Section~\ref{ssec:destructive}).
Note that, as opposed to \textit{overfitting}, overthinking leads to both wasted computation and misclassifications.
Further, whereas regularization, such as Dropout~\cite{srivastava2014dropout}, can mitigate overfitting; there is no general remedy for overthinking.

To illustrate the overthinking problem, we focus on a case study: \textit{VGG-16} network trained on Tiny Imagenet task (59\% test accuracy).
We convert this network to an SDN---to \textit{VGG-16-SDN}---via the IC-only training strategy.
The conversion allows us to monitor how the original network's predictions evolve, without altering them at all.

\subsection{The Wasteful Effect of Overthinking}
\label{ssec:wasteful}

\nstopic{Quantifying the Effect}
For \textit{VGG-16-SDN}, assuming that a sample exits at an IC right after a correct internal prediction; 24\%, 15\%, 14\%, 6\%, 8\% and 4\% of the test samples would exit at each IC.
The remaining 29\% of the samples exit at the final classifier---4\% correctly and 25\% wrongly classified.
Based on these \emph{ideal} exit rates and $\mathcal{C}_i$'s (the IC's relative inference costs); the average inference cost would decrease by 43\% as a result of eliminating overthinking.
Further, on CIFAR-10, 95\% of the samples do not require the full depth of a CNN ---81\% on CIFAR-100 and 69\% on Tiny ImageNet---; whereas the rest exit at the final classifier.
In consequence, eliminating overthinking would cut the average inference costs by 73\% on CIFAR-10, by 55\% on CIFAR-100, and by 40\% on Tiny ImageNet.
Evidently, overthinking leads to more severe waste of computation as the task gets less complex.
Note that our assumption here is impractical as it requires a perfect way to discern a wrong classification from a correct one---Section~\ref{ssec:early_exits} presents a realistic way.

\begin{figure}[tb]
    \centering
    \includegraphics[width=0.48\textwidth]{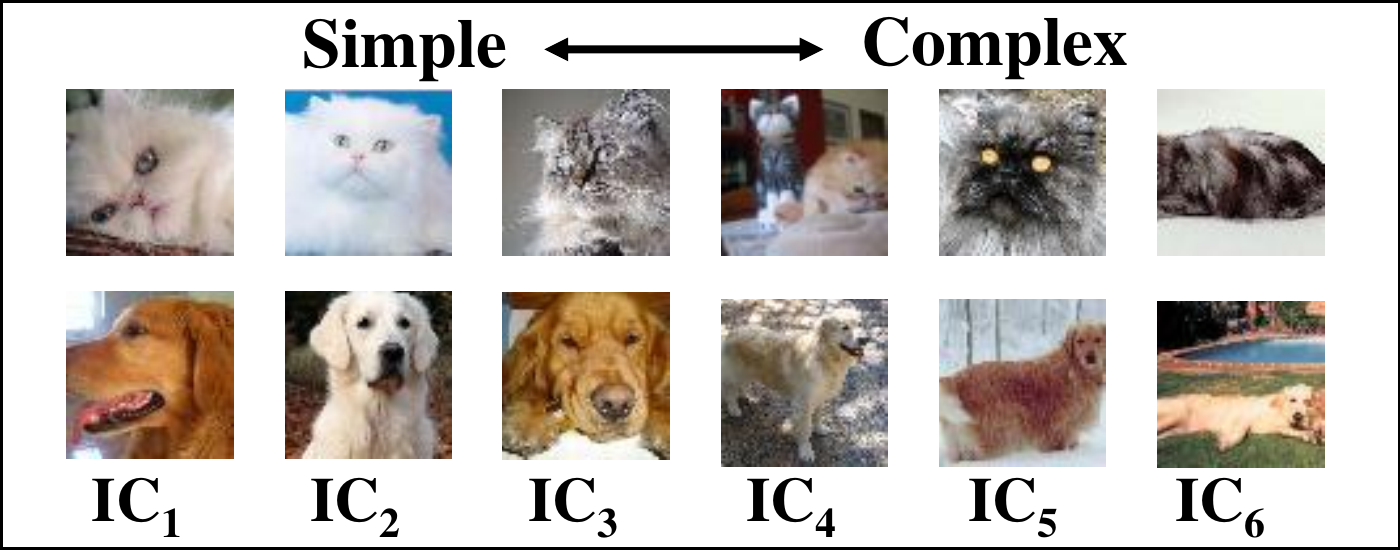}
    \caption{Sample images from the Tiny ImageNet classes \textit{Persian Cat} (top row) and \textit{Golden Retriever} (bottom row). Each column presents the samples that are correctly classified first at the given IC. We can see how the samples get progressively complex over ICs. The network is \textit{VGG-16-SDN}, trained using the IC-only training strategy.}
    \label{fig:simple_complex}
\end{figure}

\topic{Explaining the Behavior}
Considering the perfect exit rates, 29\% of the samples still need to be forwarded until the final classifier; whereas 24\% can exit at the first IC.
Figure~\ref{fig:simple_complex} shows randomly sampled images---the input samples---that exit at each IC in \textit{VGG-16-SDN}.
The first column presents the samples that are correctly classified at the first IC ($IC_1$), and so on.
In subsequent columns of the figure, notice how the input \textit{complexity} progressively increases.
These images suggest that the earlier ICs learn simple representations that allow them to recognize typical samples from a class, but that fail on atypical samples, e.g. where the subject is occluded or zoomed out.
Overthinking leads to wasted computation as the network applies unnecessarily complex representations to classify the input. 
On the other hand, not having enough representational complexity causes wrong predictions, e.g. the $IC_1$ misclassifies the samples in the remaining columns of the figure.

\subsection{The Destructive Effect of Overthinking}
\label{ssec:destructive}

\nstopic{Quantifying the Effect}
Out of 41\% of the samples \textit{VGG-16} misclassifies; 3\% are actually correctly classified first at $IC_1$ of \textit{VGG-16-SDN}---4\% at $IC_2$, 3\% at $IC_3$, 2\% at $IC_4$, 3\% at $IC_5$, and 1\% at $IC_6$.
As a result, the \textit{cumulative accuracy} of \textit{VGG-16-SDN} is 75\%---16\% higher than the original accuracy.
In cumulative accuracy, we consider a sample to be correctly classified if there is \emph{at least one} correct internal prediction---or the final prediction---of the SDN.
This metric measures the ideal accuracy a network can reach after eliminating overthinking.
The difference between the cumulative and the original accuracies quantifies the destructive effect: 4\% on CIFAR-10, 13\% on CIFAR-100, and 14\% on Tiny ImageNet task. 
This effect also explains the reason why an IC could outperform the final classifier in Table~\ref{table:internal_classifiers}: at a certain IC, the destructive effect starts to outweigh the accuracy gain from increased feature complexity; after which the accuracy starts decreasing.
Overall, the destructive effect can be seen in up to 50\% of all misclassifications a CNN makes.

\begin{figure}[tb]
    \centering
    \includegraphics[width=0.48\textwidth]{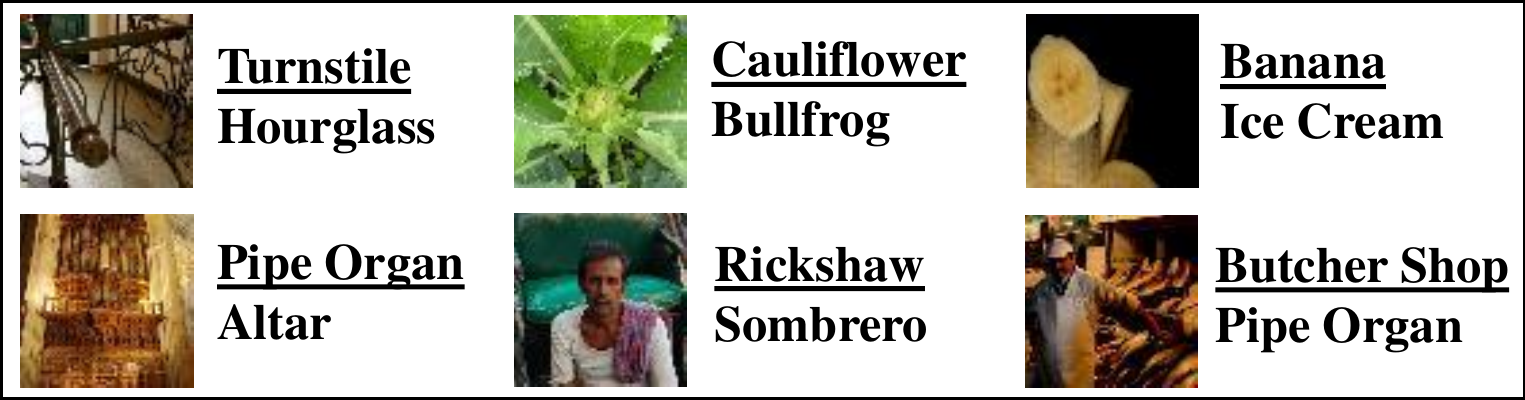}
    \caption{A set of Tiny ImageNet samples only the first internal classifier of \textit{VGG-16-SDN} can correctly classify. For each image, two labels are the correct label and the wrong prediction at the final classifier, respectively.}
    \label{fig:only_first}
\end{figure}

\topic{Explaining the Behavior}
Figure~\ref{fig:only_first} presents a selection of images that \emph{only} the first IC can correctly classify---we identified 46 of such cases. 
\textit{VGG-16} and all other ICs in \textit{VGG-16-SDN} misclassify these samples.
We hypothesize that these images consist of \textit{confusing} elements, sometimes belonging to more than one class;
while the first IC recognizes the objects displayed prominently, subsequent layers discover finer details that are irrelevant.
These results suggest that correct classifications require the \emph{appropriate} level of representational complexity for each sample, which further illustrates the utility of the ICs in an SDN. 
In Section~\ref{ssec:confusion}, building on the idea of confusion, we design a new confusion metric and visually investigate the sources of confusion in these samples.

\topic{The Destructive Effect Leads to Malicious Outcomes}
Recent work has shown the risk of backdoor attacks when the training was performed by a malicious trainer~\cite{gu2017badnets}. 
A \emph{backdoored} DNN, returned by the adversary, behaves normally on most inputs but causes targeted misclassifications when a trigger known only to the attacker is present.
Our results suggest that backdooring attacks leverage the destructive effect of overthinking.
In our experiments, we use a backdoored \textit{VGG-16} network on CIFAR-10 task.  
The trigger is a small white square at the bottom right corner, and when it is present the network classifies any input being in the \textit{dog} class.
The accuracy of the backdoored network is 92\% on clean samples and 12\% on the samples containing the backdoor---on the backdoor inputs.
The network classifies 98\% of the backdoor inputs in the dog class; which demonstrates the impact of such a threat.
Converting the backdoored network to an SDN--via the IC-only training strategy---illuminates the nature of the attack.
The attack's success before the fourth IC ($IC_4$) is minimal: $IC_4$ makes correct predictions on 87\% of the backdoor inputs.
However, at $IC_5$ and $IC_6$, the correct predictions on the backdoor inputs suddenly drop to 76\% and 38\%---and 12\% at the final classifier.
This pattern fits in the destructive effect and indicates that a technique to mitigate overthinking might mitigate the attack as well---we present these results in Section~\ref{ssec:early_exits}.

\section{Mitigating the Overthinking Problem}
\label{sec:mitigating}
After understanding the overthinking problem, in this section, we propose new remedies to mitigate its adverse effects, i.e. being wasteful and destructive.
For this purpose, we further utilize Shallow-Deep Networks as a vehicle for designing our techniques.

To prevent overthinking entirely, ideally, one could rely on early exits upon encountering a correct internal prediction.
However, this requires a perfect way to determine the correctness of a given prediction.
In human decision-making, people rely on imperfect heuristics to both conserve energy and avoid potential mistakes~\cite{fiske2013social}.
Analogously, to mitigate network overthinking, we propose two SDN-based heuristics: the confidence-based \emph{early exits} (Section~\ref{ssec:early_exits}) and network \emph{confusion} analysis (Section~\ref{ssec:confusion}).

\subsection{Confidence-Based Early Exits}
\label{ssec:early_exits}
In Section~\ref{ssec:wasteful}, we show that an ideal, but \emph{impractical}, early exit mechanism could eliminate overthinking entirely. 
Here, as a \emph{practical} mechanism, we propose using the internal prediction \textit{confidence} for determining when the network should stop thinking.
Confidence allows us to simply decide between making an early exit or forwarding the input sample to subsequent layers.
If none of the internal predictions---or the final prediction---are confident enough to for an exit; our mechanism outputs the most confident among them.
We opt for a simple confidence mechanism for highlighting that we mitigate overthinking regardless of the mechanism, which could be improved with schemes, such as~\cite{teerapittayanon2016branchynet}.
To quantify confidence, we use the estimated probability of the sample $x$ belonging to the predicted class, i.e. $\operatorname*{max}_j \mathcal{F}^{(j)}_i(x)$.
We deem a prediction confident if this probability exceeds the threshold parameter $q$.
The threshold facilitates on-the-fly adjustment of the early exits based on the resource availability and the performance requirements.
A higher $q$ value would make the early exits \emph{conservative} and, in turn, reduce the early exit rates.
In our experiments, we search for a $q$ value ($0 \leq q \leq 1$) that satisfies our computational constraints on a small unlabeled holdout set. 

\begin{table}[tb]
    \caption{Comparing the inference costs of the CNNs and SDNs with early exits. \textbf{N.} lists the original CNNs and their accuracies. \textbf{$\leq$25\%}, \textbf{$\leq$50\%}, and \textbf{$\leq$75\%} report the early exit accuracy when we limit the average inference cost to at most 25\%, 50\% and 75\% that of the original CNN's. \textbf{Max} reports the highest accuracy early exits can achieve. We highlight the cases where an SDN outperforms the original CNN. In each cell, the left and right accuracies are from the IC-only and the SDN training strategies.}
    \label{table:early_exits}
    \centering
    \begin{small}
    \begin{sc}
    \resizebox{0.48\textwidth}{!} {
    \begin{tabular}{lcccr}
        \toprule
        \textbf{N.} & \textbf{$\leq$25\%} & \textbf{$\leq$50\%} & \textbf{$\leq$75\%} & \textbf{Max} \\
        \midrule
        \multicolumn{5}{c}{\textbf{CIFAR-10}} \\
        \arch{V}\textit{(93.1)} & 84.0 $|$ 85.4 & 92.8 $|$ \textbf{93.2} & \textbf{93.2} $|$ \textbf{93.4} & \textbf{93.2} $|$ \textbf{93.4} \\
        \arch{R}\textit{(91.9)} & 57.6 $|$ 76.2 & 84.5 $|$ 91.4 & 91.4 $|$ \textbf{92.1} & \textbf{91.9} $|$ \textbf{92.1} \\
        \arch{W}\textit{(94.1)} & 67.0 $|$ 85.2 & 90.3 $|$ 93.6 & 93.5 $|$ \textbf{94.1} & 93.9 $|$ \textbf{94.1} \\
        \arch{M}\textit{(90.8)} & 87.9 $|$ 88.2 & \textbf{91.1} $|$ \textbf{91.5} & \textbf{91.3} $|$ \textbf{91.5} & \textbf{91.3} $|$ \textbf{91.5} \\
        \hline
        \multicolumn{5}{c}{\textbf{CIFAR-100}}\\
        \arch{V}\textit{(70.9)} & 53.3 $|$ 55.5 & 68.9 $|$ \textbf{72.5} & \textbf{72.5} $|$ \textbf{74.3} & \textbf{72.6} $|$ \textbf{74.4} \\
        \arch{R}\textit{(68.8)} & 46.1 $|$ 46.7 & 61.2 $|$ 65.7 & 67.2 $|$ \textbf{70.8} & \textbf{69.7} $|$ \textbf{70.9} \\
        \arch{W}\textit{(75.1)} & 50.4 $|$ 65.8 & 66.5 $|$ 74.7 & 74.3 $|$ \textbf{76.7} & \textbf{75.5} $|$ \textbf{77.3} \\
        \arch{M}\textit{(64.9)} & 55.9 $|$ 57.5 & \textbf{65.9} $|$ \textbf{68.0} & \textbf{67.5} $|$ \textbf{68.9} & \textbf{67.6} $|$ \textbf{68.9} \\
        \hline
        \multicolumn{5}{c}{\textbf{Tiny ImageNet}}\\
        \arch{V}\textit{(58.6)} & 34.4 $|$ 36.8 & 44.2 $|$ 56.2 & \textbf{58.5} $|$ \textbf{63.1} & \textbf{60.4} $|$ \textbf{63.4} \\
        \arch{R}\textit{(53.9)} & 23.9 $|$ 39.0 & 37.6 $|$ 39.1 & 49.2 $|$ 52.8 & \textbf{54.0} $|$ \textbf{55.1} \\
        \arch{W}\textit{(60.3)} & 26.7 $|$ 36.6 & 42.5 $|$ 54.9 & 56.1 $|$ \textbf{62.6} & \textbf{61.0} $|$ \textbf{62.8} \\
        \arch{M}\textit{(59.3)} & 36.0 $|$ 45.3 & 55.5 $|$ 57.3 & 59.1 $|$ \textbf{61.5} & \textbf{60.3} $|$ \textbf{61.8} \\
        \bottomrule
    \end{tabular}
    }
    \end{sc}
    \end{small}
\end{table}
\raggedbottom

\topic{Early Exits Mitigate the Wasteful Effect}
Table~\ref{table:early_exits} presents the early exit accuracy of our SDNs with a limited computational budget, in two training strategies.
We calculate the computational cost when the input samples are evaluated individually; similar to previous studies~\cite{teerapittayanon2016branchynet, huang2017multi}.
We measure the accuracy under three budget settings: not exceeding 25\%, 50\% and 75\% of the original CNN's inference cost.
We control an SDN's average inference cost by adjusting the threshold parameter $q$.
We also report the highest accuracy early exits can achieve, without any specific budget. 
First, the SDN training improves the early exit accuracy significantly; exceeding the original accuracy while reducing the inference costs more than 50\% on CIFAR-10 and CIFAR-100 tasks. 
Even with the IC-only training strategy, early exits are still effective; allowing more than 25\% reduced inference cost.
In our most complex task, Tiny ImageNet, early exits can reduce the cost by more than 25\%, usually without any accuracy loss.
Overall, an SDN's early exits can mitigate the wasteful effect of overthinking.

\topic{Early Exits Mitigate the Backdoor Attack}
In Section~\ref{ssec:destructive}, we identified that a backdooring attack on \textit{VGG-16} induces the destructive effect.
Our early exit mechanism significantly reduces the success of this attack.
When the threshold is at $q = 0.8$; the backdoored network makes correct predictions on 84\% of the backdoor inputs---up from 12\% without early exits.
Further, the network classifies only 17\% of the backdoor samples to the attacker's target class---down from 98\%.
Overall, our results suggest that early exits can mitigate this attack.
We believe that Shallow-Deep Networks shed light on potential avenues for a defensive strategy against backdooring attacks.

\topic{The Destructive Effect Causes Low Confidence}
We see that early exits increase the accuracy by up to 8\% over the original CNN; especially on more difficult tasks.
This hints that early exits, to a certain extent, can recover the accuracy lost to the destructive effect.
However, the samples that trigger the destructive effect receive significantly lower correct prediction confidence--- $\sim$0.3 on average vs. $\sim$0.6 in other correct predictions. 
These low confidence samples, as a result, pass up the early exits; even though they are correctly classified.
The disagreements among internal predictions on these samples indicate that the network is \textit{confused} while the predictions are evolving. 
The confusion reduces the benefit from early exits; however, it could also provide valuable insights into how the network reaches a prediction.
Next, we explore the notion of network confusion further.

\subsection{Network Confusion Analysis}
\label{ssec:confusion}
An SDN's internal predictions reveal how consistently the network reaches its final prediction.
Disagreements among them hint that the prediction is inconsistent and \textit{confused}; whereas an agreement indicates consistency.
The destructive effect of overthinking also displays a pattern of disagreement---$\hat{y}_i = y \neq \hat{y}_{final}$---, and confusion.
We propose the \textit{confusion} metric to capture this inconsistency.
The confusion metric quantifies how much the final prediction diverged from the internal predictions.
The divergence of the final prediction from an internal prediction is given by the $L_1$ distance between them, i.e. $\mathcal{D}_i(x) = ||\mathcal{F}_{final}(x) - \mathcal{F}_i(x) ||_1$. 
We observe that, compared with other distance metrics, $L_1$ distance is more consistent for capturing the small differences between two predictions.
The summation over all internal predictions, i.e. $\sum_{\forall i} D_i(x)$, gives the unbounded confusion score on the sample $x$.
We normalize the unbounded scores using the mean and the standard variation of the training set samples.

\begin{figure}[tb]
    \centering
    \includegraphics[width=0.48\textwidth]{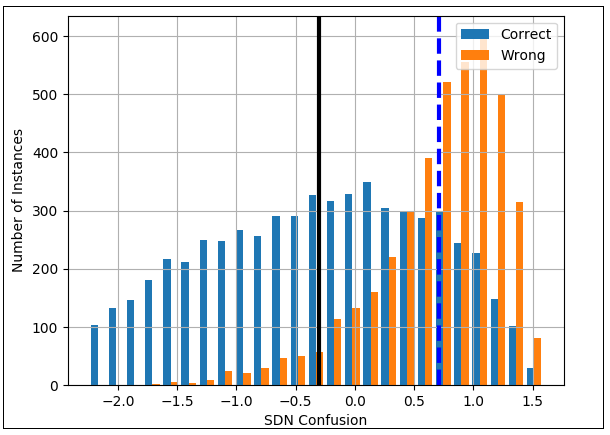}
    \caption{The distribution of the confusion scores \textit{VGG-16-SDN} produces on Tiny ImageNet test samples. The dotted and straight lines indicate the average confusion scores in the wrong and correct predictions, respectively.}
    \label{fig:confusion_scores}
\end{figure}

\topic{Confusion Metric is an Error Indicator}
Section~\ref{ssec:destructive} demonstrates the prevalence of the destructive effect on misclassifications.
The confusion metric inherently captures this effect.
As a result, we expect confusion scores to be a reliable indicator for the cases when the network misclassifies a sample.
Such indications have practical significance for handling errors; for example, they can alert users about cases where the network is unable to make a good prediction. 
Here, we continue our case study on \textit{VGG-16} and \textit{VGG-16-SDN} on Tiny ImageNet task.
Figure~\ref{fig:confusion_scores} presents \textit{VGG-16-SDN}'s normalized confusion scores in correct and wrong final predictions on test samples.
While correct predictions tend to have low confusion scores ($-0.29$ on average), the misclassifications are concentrated among instances with high confusion ($0.71$ on average)---more than one standard deviation difference.
As a baseline, we consider the prediction confidence scores, i.e. $\operatorname*{max}_j \mathcal{F}_{final}^{(j)}(x)$, \textit{VGG-16} produces.
The difference in confidence scores is less pronounced: $0.93$ vs. $0.71$ on the correct and wrong predictions, respectively---less than one standard deviation difference.
Further, when used as an indicator for likely misclassifications, confusion also produces fewer false negatives than confidence.
Compared to an average correct prediction, 5.5\% of the misclassified instances (228 out of 4135) cause less confusion for the SDN; whereas 24\% (980 out of 4135) misclassified instances obtain more confidence in the original CNN.
Altogether, the SDN confusion metric is a reliable and non-expensive indicator of whether a misclassification is likely.

\begin{figure}[tb]
    \centering
    \includegraphics[width=0.48\textwidth]{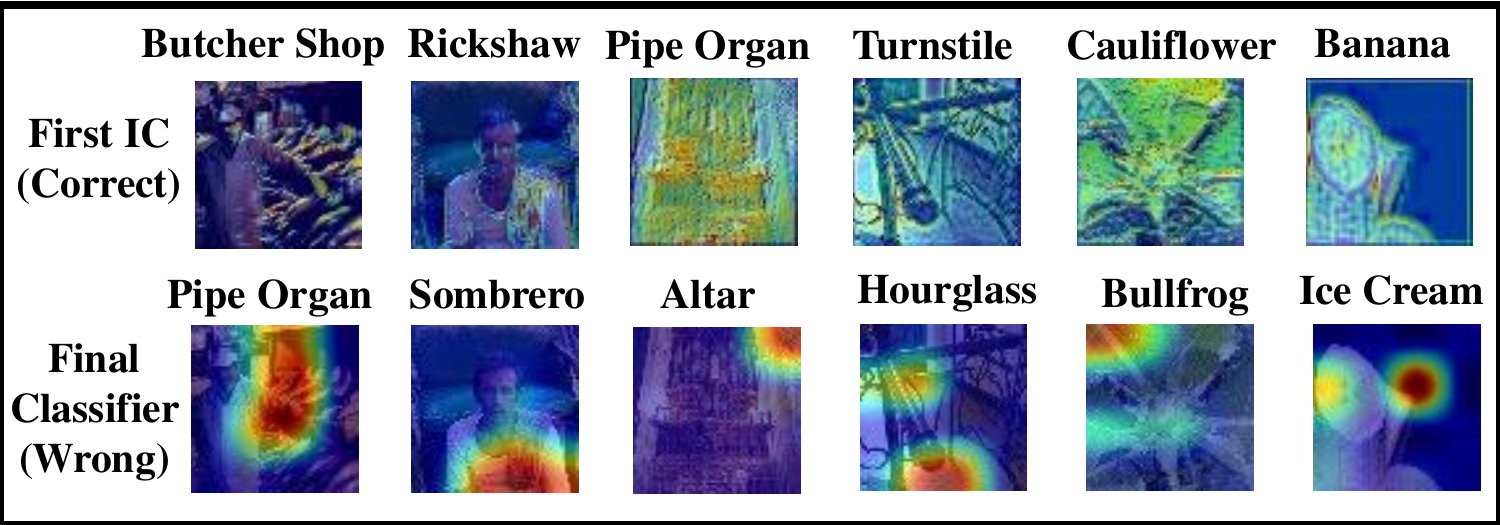}
    \caption{The visual interpretation of network confusion. Green color indicates the elements influential in the prediction, and blue color signifies less influence.}
    \label{fig:confusion}
\end{figure}

\topic{Visualizing Confusion Helps with Error Diagnosis}
Towards error mitigation, our confusion metric can indicate whether the network is likely to make a mistake.
As a further step, we propose visualizing the confusion for investigating sources of the destructive effect in the input.
We hypothesize that the confusing elements in an input trigger disagreements and misclassifications.
Specifically, given an input $(x,y)$, we visualize the wrong final prediction and the disagreeing $i^{th}$ correct internal prediction, i.e. $\hat{y}_i = y \neq \hat{y}_{final}$.
We use Grad-CAM~\cite{selvaraju2017grad} for visualizing the input elements that are influential in a prediction.
In Figure~\ref{fig:only_first}, we presented some images that effectively confuse \textit{VGG-16-SDN}.
On these confusing images, the first IC, which predicts the correct labels, and the final classifier, which misclassifies, disagree.
Figure~\ref{fig:confusion} shows the Grad-CAM visualizations of the first IC and final classifier, on these images.
We see that the first IC focuses on the general structure of the input, whereas the final classifier focuses on the fine details.
As an example, consider the \textit{banana} vs. \textit{ice cream} image. 
Here, the first internal classifier focuses on the simple structure of the banana, which leads to a correct prediction.
The confusing details, such as the exposed top part of the banana, mislead the final classifier into classifying the image as ice cream.
Visualizing the confusion helps us to better understand how the network makes incorrect decisions; providing a new perspective towards interpretable deep learning.

%
%

\section{Conclusions}
\label{sec:conclusions}
We introduce Shallow-Deep Networks (SDNs), a modification for obtaining accurate predictions from the internal layers of an off-the-shelf deep neural network (DNN).
Our modification allows us to monitor how predictions evolve throughout the forward pass and provides a new way to understand DNNs.
SDNs help us expose the prevalence of the \emph{overthinking} problem in convolutional neural networks, which leads to wasted computation and to misclassifications on both natural and adversarial inputs. 
SDNs also enable new mechanisms for mitigating the problem.
Confidence-based early exits significantly reduce the average inference cost while preserving the original performance.
%
%
Moreover, our \emph{confusion} analysis exposes disagreements among internal classifiers
and reliably indicates the cases where the network is likely to misclassify an input.
This paves the way for visually interpreting errors by revealing the input elements responsible for the confusion. 
We hope that our findings lay the foundations for more efficient and more accurate networks that are not susceptible to the prevalent overthinking problem.
\section*{Acknowledgements}
\label{sec:ack}
We would like to thank Dr. Tom Goldstein, Dr. Nicolas Papernot, Virinchi Srinivas and Sarah Joseph for their valuable feedback. This research was partially supported by the Department of Defense.

\bibliographystyle{icml2019}
\bibliography{refs}

\end{document}